\documentclass[conference]{IEEEtran}
\IEEEoverridecommandlockouts
\usepackage{cite}
\usepackage{amsmath,amssymb,amsfonts}
\usepackage{textcomp}
\usepackage{xcolor}
\usepackage[numbers,sort]{natbib}
\usepackage{color}
\usepackage{xcolor}
\newcommand{\minisection}[1]{\vspace{0.025in} \noindent {\bf #1}\ }

\usepackage{gensymb}
\usepackage{algorithm}
\usepackage{algpseudocode}
\usepackage{multicol}
\usepackage{caption}
\usepackage{subcaption}
\usepackage{graphicx}
\usepackage{bm}

\def\BibTeX{{\rm B\kern-.05em{\sc i\kern-.025em b}\kern-.08em
    T\kern-.1667em\lower.7ex\hbox{E}\kern-.125emX}}
\begin{document}

\title{Improving Generalization in Game Agents with Data Augmentation in Imitation Learning}

\author{
Derek Yadgaroff$^{1}$,
Alessandro Sestini$^{2}$,
Konrad Tollmar$^{2}$,
Ayca Ozcelikkale$^1$,
Linus Gisslén$^{2}$
\\
$^{1}$\textit{Uppsala University}, $^{2}$\textit{SEED - Electronic Arts (EA)}
\\
derekyadgaroff@gmail.com,  ayca.ozcelikkale@angstrom.uu.se, \{asestini, ktollmar, lgisslen\}@ea.com
}

% \author{\IEEEauthorblockN{Anonymous Authors}}

\maketitle

\begin{abstract}
% Improved abstract - please use if we are allowed to change it
Imitation learning is an effective approach for training game-playing agents and, consequently, for efficient game production. However, generalization---the ability to perform well in related but unseen scenarios---is an essential requirement that remains an unsolved challenge for game AI. Generalization is difficult for imitation learning agents because it requires the algorithm to take meaningful actions outside of the training distribution. 
In this paper we propose a solution to this challenge. Inspired by the success of data augmentation in supervised learning, we augment the training data so the distribution of states and actions in the dataset better represents the real state--action distribution.
This study evaluates methods for combining and applying data augmentations to observations, to improve generalization of imitation learning agents. It also provides a performance benchmark of these augmentations across several 3D environments. 
These results demonstrate that data augmentation is a promising framework for improving generalization in imitation learning agents.

%Imitation learning is an effective approach for training game-playing agents and subsequently, for efficient game production. However, generalization, the ability to perform well in related but unseen scenarios, is an essential requirement that remains an unsolved challenge for game AI. This problem is difficult for imitation learning agents as it requires the algorithm to take meaningful actions outside of the training distribution. 
%In this paper we propose a solution to this challenge. Inspired by the success of data augmentation in supervised learning, we augment the training data so that the distribution of states and actions in the dataset better represent the real state-action distribution.
%We evaluate ways to combine and use data augmentations on observations for improving generalization of imitation learning agents. This study provides a performance benchmark of these augmentations across a number of 3D environments. 
% These results illustrate that data augmentation is a promising framework for improving generalization.

\end{abstract}


\section{Introduction}
Playtesting is an essential component of modern video game production. Gameplay issues and bugs can significantly reduce the quality of the user experience. The process of game testing is usually done by designated human testers, both internal and external to the development team. However, due to the increased size and complexity of modern video games, manual playtesting is becoming expensive in regard to time, manpower, resources, and budget. Stakeholders in academia and industry have recently proposed automated testing solutions that could free up human resources to focus on more meaningful tasks, such as evaluating gameplay balance, difficulty, and player engagement.

Current automated testing techniques rely primarily on model-based automated playtesting~\cite{playfulness, scenariobased}, but recent developments in Reinforcement Learning (RL)~\cite{alphastar,openai,gtsophy} have demonstrated that it is possible to train game Artificial Intelligence (AI) agents to reach human-level performance in complex video games, requiring however a vast amount of resources to achieve such training. These techniques can also train agents to explore game scenes and exploit environments to detect bugs~\cite{augmenting, improving}, thereby improving automated playtesting.

However, most techniques used for automated game testing suffer from an efficiency--generalization trade-off. Model-based and RL policies are known to easily overfit~\cite{sestini2022kiwi,illuminating}, and achieving greater generalization requires numerous training samples. Imitation Learning (IL) was proposed to reduce the sample-inefficiency problem of RL~\cite{sestini2022ccpt, sestini2022kiwi}. An IL agent learns from demonstrations rather than experience, allowing game designers to use prior knowledge to guide the agent towards its goal without additional knowledge of programming or machine learning. Even though IL can mitigate the sample-inefficiency problem, the generalization issue persists.
To develop an IL agent that can address situations beyond the demonstration set and minimize the distributional shift issue----the phenomenon where the distribution of states and actions encountered by the learning agent differs from those in the expert demonstration data, leading to sub-optimal performance---a significant number of datasamples are needed. However, producing a large demonstration set can be costly. Additionally, even with a considerable amount of data, IL might still overfit to the training environment. Ideally, we should test the IL policy across various versions of the game, simulating game developers modifying it. Incorporating priors for new environments may be challenging, as developers may not anticipate future changes. Hence, it is crucial to have a model that can generalize beyond the training environment. 

In this paper, we investigate how to improve the generalization of IL agents by reducing sample-inefficiency via data augmentation. While data augmentation is commonly used to address generalization and efficiency in computer vision problems~\cite{supervisedaug1, supervisedaug2}, recent applications of image-based~\cite{drq} and feature-based~\cite{s4rl} state spaces have shown promising results in both online and offline RL. Consequently, we propose a comprehensive study of the different data augmentation techniques in IL, with a focus on augmented feature-based state spaces, as game AI typically operates with feature values instead of images. We address the following research questions: Which augmentation techniques perform best in a game AI IL setting? Which consistently perform well in different use cases? And which are ineffective in a game AI IL setting? To answer these questions, we train agents with different augmentation combinations in a training environment and evaluate their generalization across four test environments. We do not propose any new data augmentation techniques; instead, we conduct a comprehensive study of existing ones used in various domains and present an analysis of the combination of augmentations that yield the best results. This study shows that certain combinations are more suitable for this setting than others, with the best achieving a performance $1.6$ times higher than that of the non-augmented agent.

Our main contribution aims to provide guidance to those seeking to use data augmentation to increase generalization in IL agents, particularly in the game testing context. It can also be used as a potential starting point for improving generalization of any feature-based game AI agent.

\section{Related Work}
While generalization has been widely discussed in reinforcement learning literature \cite{illuminating, procgen}, only a few works have focused on generalization and imitation learning. In this section, we review the most relevant literature related to this work.

\subsection{Imitation Learning and Games}
Imitation learning aims to distill, from a dataset of demonstrations, a policy that mimics the behavior of an expert demonstrator. It is assumed that the demonstrations come from an expert exhibiting near-optimal behavior. Standard approaches are based on Behavioral Cloning (BC), and mainly employ supervised learning~\cite{bc1,dagger}. Generative Adversarial Imitation Learning (GAIL) is a recent IL technique based on a generator--discriminator approach~\cite{gail}.

IL has emerged as a widely accepted method for creating agents to perform tasks in 3D video games without RL~\cite{sestini2022kiwi, ilcounter}. \citet{minecraft} used an agent trained solely through BC to play the game Minecraft. \citet{reveal} used demonstrations to guide exploration, while \citet{winning} employed a BC approach to teach agents based on game designers' expertise. \citet{concurrent} trained agents using a combination of IL and RL with multi-action policies for a first person shooter game. Additionally, \citet{adam} used an inverse RL technique to train agents capable of playing a variety of games in the Arcade Learning Environment suite \cite{ale}. Other notable examples include: the work of \citet{sestini2022ccpt}, who combined RL, IL, and curiosity-driven learning to train playtesting agents; the work of \citet{ferguson2022imitating}, who proposed the use of dynamic time warping IL to imitate distinct playstyles; and the work of \citet{imitating}, who introduce several innovations to make diffusion models suitable for imitating human behaviors in a first-person shooter game. However, all of these models must address the efficiency--generalization trade-off: higher levels of generalization require more demonstrations or more interactions with the environment. In this paper we examine the latest data augmentation techniques used in RL to increase generalization and efficiency and apply them in IL.

\subsection{Generalization in Reinforcement Learning}
In several use cases, policies produced with RL, IL or inverse RL have overfit to their training environments~\cite{procgen, deairl}. Techniques like domain randomization and procedural content generation improve generalization and help prevent overfitting~\cite{deepcrawl, teachmyagent}. In inverse RL, \citet{deairl} proved that with the correct training setup, domain randomization can mitigate overfitting and improve efficiency. However, domain randomization requires having a procedurally ready game or access to the environment code to implement randomized elements. In practical settings such as video game production, domain randomization is not always feasible.

Data augmentation has also been investigated in the context of RL and generalization~\cite{survey}. \citet{rlaugmentation} proposed a thorough survey regarding data augmentation and RL for image data, while \citet{augmentationil} combined data augmentation and the GAIL algorithm to train a sample-efficient IL agent in a robotic setting. The present work takes inspiration from the S4RL algorithm of \citet{s4rl}. These authors investigated the efficacy of using data augmentation in offline reinforcement learning on a feature-based state space. They proposed $7$ different augmentation schemes and analyzed their performance with existing offline reinforcement learning algorithms. Furthermore, they combined the most successful data augmentation scheme with a state-of-the-art offline RL technique. Similarly, this present work investigates the effects of $6$ feature-based data augmentation techniques for training game-playing agents with IL.

\section{Method}
\label{sec:method}

Here we first review the main IL technique, behavioral cloning, and discuss the rationale behind its selection. Second, we review data augmentations previously studied in offline RL research and explain how we adapt these techniques to improve sample efficiency and generalization in our use case. 

\subsection{Behavioral Cloning}
\label{bc}
We begin by introducing the main algorithm used in this study. Since the focus is on data augmentation rather than selecting the optimal algorithm, we choose one of the simplest IL algorithms, BC. Recent literature has shown BC to be a good algorithm for IL in 3D games~\cite{sestini2022kiwi, ilcounter}. Formally, we describe our domain as a Markov Decision Process (MDP) consisting of a state space $S \in \mathbb{R}^n$, an action space $A \in \mathbb{R}^m$, a transition dynamic function $p(s'|s,a): S \times A \rightarrow p(S)$ and a parameterized policy $\pi_\theta:S \rightarrow p(A)$ that is a mapping from the state space $S$ to a probability distribution over the set of actions $A$. Section~\ref{sec:setup} provides a complete description of the particular state- and action-space used in this work. Note that in our setting we do not have a reward function since the agent is learning from expert demonstrations. Given a set of optimal demonstrations $\mathcal{D} = \{\tau_i \; | \; \tau_i = (s_0^i, a_0^i, ..., s_{T_i}^i, a_{T_i}^i), \; i=1, .., N\}$ of $N$ trajectories $\tau_i$, each composed of a sequence of state-action pairs $(s^i_k, a^i_k)$, we define the BC objective used to update the network as:
\begin{equation}
\label{eq:bc}
    \arg\max_{\theta} \mathbb{E}_{(s,a) \sim \mathcal{D}}[\log \pi_\theta(a | s)].
\end{equation}
This equation follows the maximum entropy objective: by increasing the log-probability of the policy $\pi(a_t|s_t)$ for a given $(a_t, s_t) \sim D$ pair sampled from the expert dataset $\mathcal{D}$, we enhance the probability of sampling that action in that specific state or similar states in proximity to $s_t$. At optimality, the policy $\pi$ mimics the expert behavior represented by $\mathcal{D}$.

\subsection{Data Augmentations}
\label{augmentation.list}
Inspired by the success of data augmentation for computer vision and online and offline RL, we apply a set of data augmentation techniques to our data. As we previously mentioned, our dataset consists of $(s, a)$ pairs, but we augment only the state $s$ and retain the original action $a$ for the augmented state. To accomplish this, as noted by \citet{s4rl}, we assume that applying a small transformation to an input state $s$ results in a physically realizable state $\hat{s}$, and that the original action $a$ remains a valid action for the state $\hat{s}$. For the experiments detailed in Section~\ref{sec:evaluation}, to ensure this assumption holds, we use our prior knowledge of the environment to define our augmentations and select their hyperparameters.  

We denote a data augmentation transformation as $\tau_i(\hat{s_{t}}|s_{t})$, where $s_{t} \sim D$ is an original state sampled from the dataset $D$ at timestep $t$, $\hat{s_{t}}$ is an augmented state, and $\tau_i \in \mathcal{T}$ is a stochastic augmentation with $\mathcal{T}$ representing the set of all available augmentations defined below. 
In contrast to \citet{s4rl}, who used only $1$ augmentation per dataset, we apply from $1$ to $3$ sequential augmentations (see Section~\ref{sec:evaluation}). In this way we assess whether a combination of augmentations performs better than individual ones. For instance, if we augment the data with $\tau_1$, $\tau_2$, and $\tau_3$, the resulting state will be $\hat{s_t} = \tau_{3}(\tau_{2}(\tau_{1}(s_{t})))$. 

As we see in Section~\ref{section:environment}, our state space is composed of both continuous and categorical values, and we choose to keep these separate when applying our augmentations. 
In particular, Gaussian noise, uniform noise, scaling, state mix-up and continuous drop-out are only applied to the continuous values of the state vector whereas semantic dropout is only applied to the categorical values.  
With that in mind, we next define the set of augmentations $\mathcal{T}$. 
For the sake of convenience, we represent vectors with i.i.d. components with a certain distribution with $s \sim X()$, e.g. $s \sim N(\mu,\sigma )$ represents a vector of i.i.d. Gaussian variables with mean $\mu$ and standard deviation $\sigma$. 

% hyperparams

\minisection{Gaussian Noise:} We sample $\epsilon \sim N(\mu,\sigma )$ and let \mbox{$\hat{s}_t = \tau_{\text{gauss}}(s_{t}) = s_{t} + \epsilon$}. We set $\mu = 0$ and experiment with different values of $\sigma$ from $\{0.03, 0.003, 0.0003, 0.00003\}$.

\minisection{Uniform Noise:} We sample $\epsilon \sim U(-\lambda, \lambda)$ and let \mbox{$\hat{s}_t = \tau_{\text{uni}}(s_{t}) = s_{t} + \epsilon$}. We set $\lambda = 0.0003$. 

\minisection{Scaling:} We sample $\epsilon \sim U(\alpha, \beta)$ and let $\hat{s}_t = \tau_{\text{sm}}(s_{t}) = s_{t} * \epsilon$ where $*$ represents element-wise multiplication. We set $\alpha = 0.0003$ and $\beta = 0.0006$. \cite{rlaugmentation}.

\minisection{State-mixup:} We sample $\epsilon \sim \beta(\alpha, \alpha)$ with $\alpha = 0.4$ and let \mbox{$\hat{s}_t = \tau_{mixup}(s_{t}) = \epsilon*s_{t} + (1-\epsilon)*s_{t+1}$} \cite{zhang2017mixup}.

% \minisection{Continuous Dropout:} We zero out some random values of the \textit{continuous} set of values in $s_t$. In particular, let $\epsilon$ be a vector of $1$'s with the same shape as the continuous set of the state $s^c_{t}$. Sample $n$ random, valid, and unique, indices from $\epsilon$ and set those elements to 0. The resulting state is defined as $\hat{s}_t = \tau_{\text{drc}}(s_{t}) = s_{t} * \epsilon$. For our experiments, we use $n=3$.

\minisection{Continuous Dropout:} We zero-out values of randomly chosen subset of continuous-valued elements of $s_t$. In particular, let $\epsilon$ be a vector of $1$'s with the same size as $s_t$.  Let $\mathcal{S}$ be the indices of the elements of  continuous-valued part of $s_t$. We sample $n$ random and unique indices from $\mathcal{S}$ and set the elements of $\epsilon$ corresponding to these indices to zero. The resulting state is defined as $\hat{s}_t = \tau_{\text{drc}}(s_{t}) = s_{t} * \epsilon$. For our experiments, we set $n=3$.

\minisection{Semantic Dropout:} We zero-out some random values of the \textit{categorical} set of values in the state $s_t$. The procedure is the same as for continuous dropout, but we use $n = 12$. The resulting state is defined as $\hat{s}_t = \tau_{\text{drs}}(s_{t}) = s_{t} * \epsilon$.

\section{Experimental Setup}
\label{sec:setup}
Here we detail the main components used in our evaluation, including the environment, the state space, the neural network architecture, and the training setup.

\subsection{Environment and State Space}
\label{section:environment}
The environment used in this study is the same as used by \citet{sestini2022kiwi}. The game is an open-world city simulation, as illustrated in Figure \ref{fig:environments}(a) and Figure~\ref{fig:environments}(b). The agent has a discrete action space of size $9$, consisting of moving forward, moving backward, right rotation, left rotation, jumping, shooting, right sidestep, left sidestep, and no action. In our experiments, as we eliminate the presence of enemies from the environment, we do not utilize the shooting action. The state space is the same proposed by \citet{sestini2022kiwi} and consists of a goal position, represented as the $\mathbb{R}^2$ projections of the agent-to-goal vector onto the $XY$ and $XZ$ planes, normalized by the gameplay area size, along with game-specific state observations such as the agent's climbing status, contact with the ground, presence in an elevator, jump cooldown, and weapon magazine status. Observations includes a list of entities and game objects that the agent should be aware of, e.g., intermediate goals, dynamic objects, enemies, and other assets that could be useful for achieving the final goal. For these entities, the same relative information from agent-to-goal is referenced, except as agent-to-entity. For our use case, the entities are the button to open the door and the goal itself. Plus, a 3D semantic map is used for local perception. This map is a categorical discretization of the space  around the agent. Each voxel in the map carries a semantic integer value describing the type of object at the corresponding game world position. We use a semantic map of size $5 \times 5 \times 5$. In this environment, an episode consists of a maximum of $750$ steps. An episode is marked a success if the agent reaches the goal before the timeout. As we discuss in Section~\ref{sec:evaluation}, from this game we create a training environment and different testing ones.

\subsection{Neural Network}
\label{section:nn}
We use the same neural network as in the work by \citet{sestini2022kiwi}. First, all the information about the agent and a goal is fed into a linear layer with ReLU activation, producing the self-embedding $x_\text{a} \in \mathbb{R}^{d}$, with $d=128$. The list of entities is passed through a separate linear layer with shared weights, producing embeddings $x_{\text{e}_i} \in \mathbb{R}^{d}$, one for each entity $\text{e}_i$ in the list. Each embedding vector was concatenated with the self-embedding, producing $x_{\text{ae}_i} = [x_\text{a}, x_{\text{e}_i}]$, with $x_{\text{ae}_i} \in \mathbb{R}^{2d}$. The list of vectors is then input to a transformer encoder with $4$ heads and average pooling, producing a single vector $x_\text{t} \in \mathbb{R}^{2d}$. In parallel, the semantic occupancy map $M \in \mathbb{R}^{5 \times 5 \times 5}$ is first input into an embedding layer of size 8 with $\tanh$ activation and then into a 3D convolutional network with three convolutional layers, with 32, 64, and 128 filters, stride 2, and leaky ReLU activation. The output of the latter component is a vector embedding $x_\text{M} \in \mathbb{R}^{d}$ that is concatenated with $x_\text{t}$, producing $x_\text{Mt}$. We then input $x_\text{Mt}$ through one linear layer of size 256 and ReLU activation, and one final linear layer of size $9$ and softmax activation, resulting in the action probability distribution.

\subsection{Training Setup}
Training is a supervised learning problem where the input data for each model is the unique dataset described in Section~\ref{section:augmented.data}. We use a neural network, as described in Section~\ref{section:nn}, built with the TensorFlow framework. Each model is trained for $300$ epochs on a single machine with an AMD Ryzen Threadripper 1950X 16-Core Processor and an NVIDIA GV100 GPU, with an average training time of $336$ seconds. 

\section{Experiments and Results}
To evaluate the effect of data augmentation on an IL agent's generalization performance, we performed an exhaustive study of unique combinations of different data augmentations. We first collected a set of demonstrations using a training environment as described in Section~\ref{section:environment}. We then trained a model for each augmentation combination using this set. Subsequently, we tested our augmented models in four different test environments with modified designs compared to the training environment. We assessed the augmented models against each other and against a baseline model, which was trained only with the original dataset without augmentation.

\begin{figure}
    \centering
    \begin{tabular}{c}
         \includegraphics[width=.71\columnwidth]{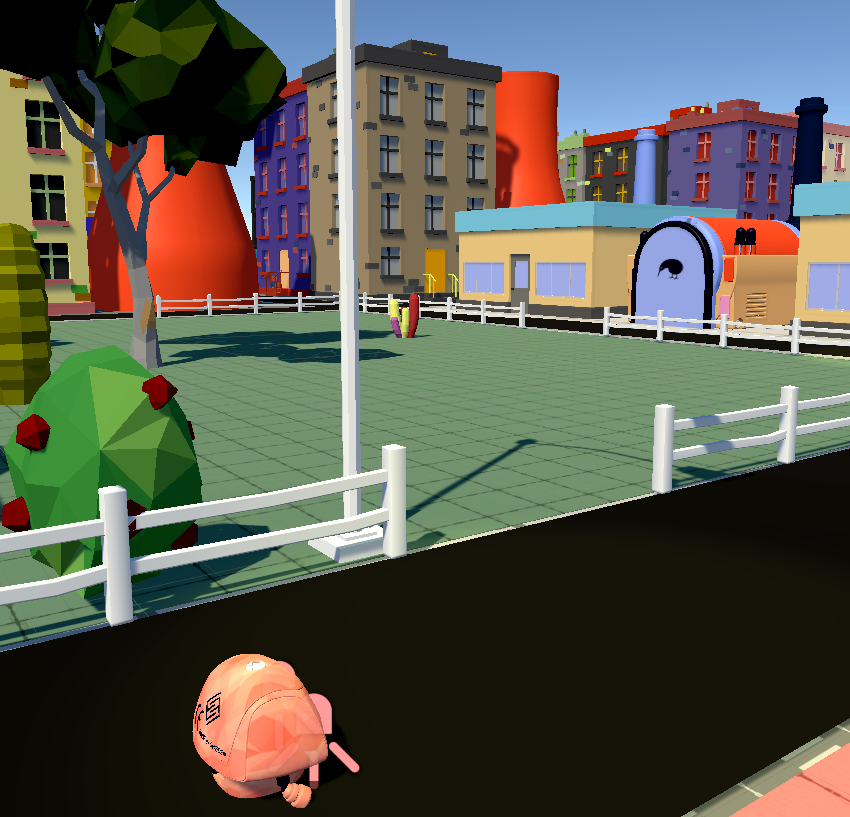} \\
         (a) \\
         \includegraphics[width=.71\columnwidth]{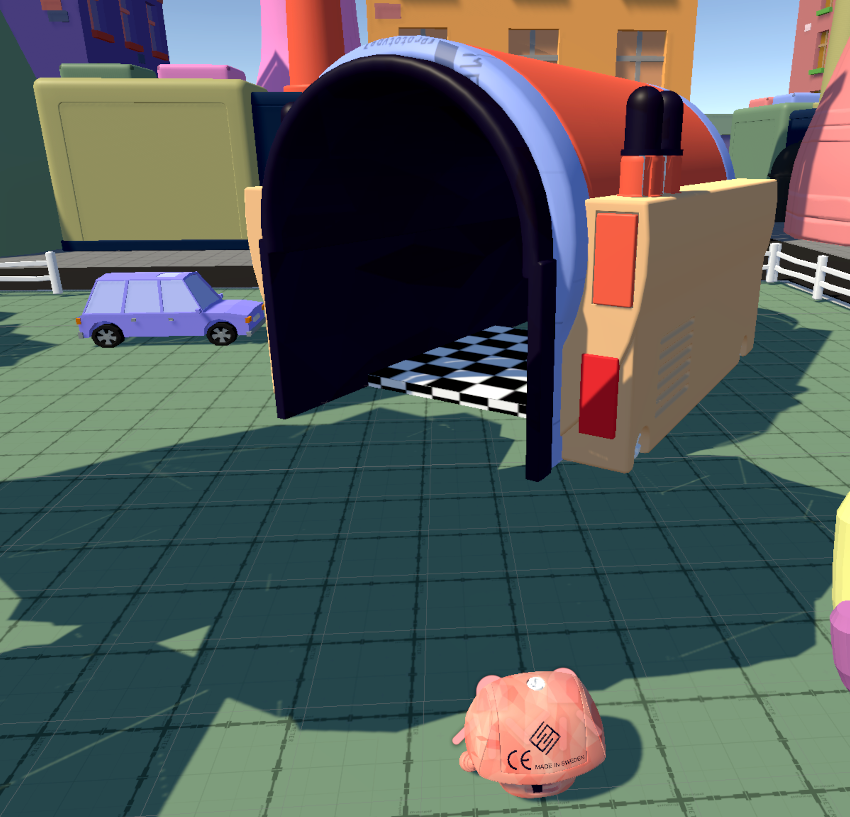} \\
         (b) \\
          \includegraphics[width=.71\columnwidth]{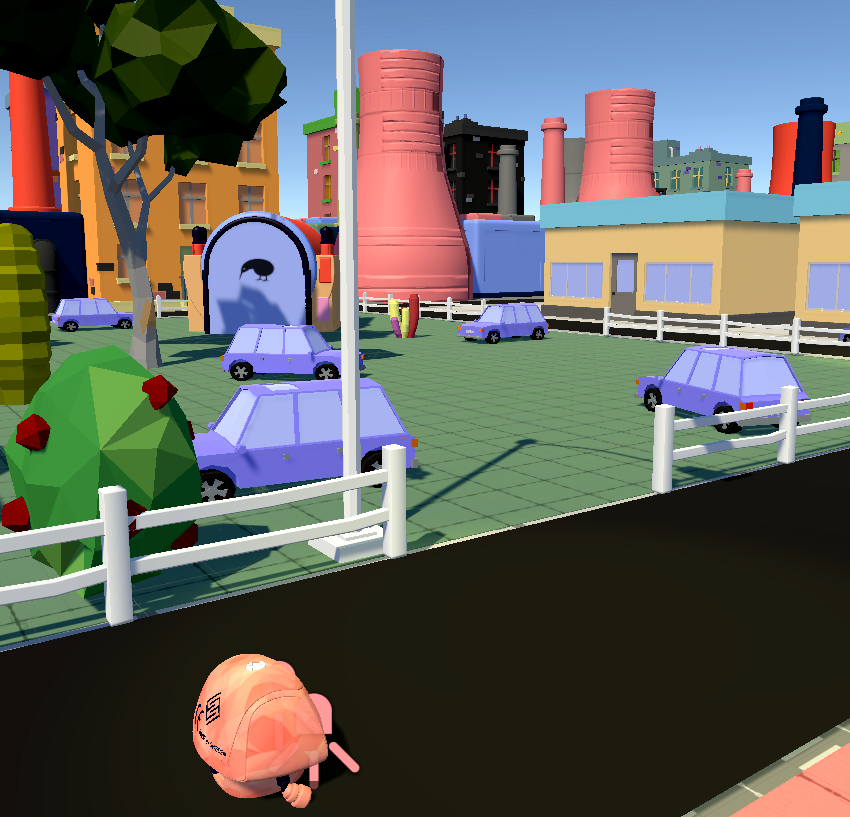} \\
         (c)
          
    \end{tabular}
    \caption{(a) The original environment where human demonstrations were created and used for training models. (b) The player/agent must navigate to the building, press the button to open the door, and enter the goal state before the door closes. (c) A modified version of the original environment used for testing the model's performance and ability to generalize in a new environment.}
    \label{fig:environments}
\end{figure}

In the training environment, we began by creating a set of demonstrations presenting the agent with the desired goal: the agent must reach a building, press a button to open a door, and enter the building before the door closes in order to reach the goal position. This task is the same Use Case 1 in the work by \citet{sestini2022kiwi}. The complexity was relatively low, with few obstacles between the agent's starting position and the goal. For the human player, there were two natural paths to reach the goal: diagonally through the park, or down the road and to the left. We used both paths to increase the range of possible states that the agent could encounter. We alternated between these two paths to maintain a balanced dataset for supervised learning. We provided human demonstrations totaling $78$ episodes, or $15,380$ samples.

For the test environments, we created four distinct new environments from the initial training environment by making design changes to the original. This included moving and rotating the goal building, increasing the number of obstacles, and obstructing one of the direct paths to the goal building. For each environment, the agent had to complete the same task of navigating to the building, opening the door, and entering the goal position. 
Herein, we describe the list of modifications for each test environment:

\begin{table*}
\caption{An overview of the changes made to the training environment to create the $4$ distinct test environments. These changes were implemented to evaluate the agent's generalization performance.
}
\begin{center}
\scalebox{0.9}{
\begin{tabular}{|c|c|c|c|}
\hline
\textbf{\textit{Environment}}& \textbf{\textit{Goal Position}}& \textbf{\textit{Obstacles}}& \textbf{\textit{Subjective Difficulty}} \\
\hline
Original (Train) & x=$0$, $z=0$, $rot=0\degree$ & $1$ obstacle  & easy \\
\hline
Test 1 & $x=2$, $z=8$, $rot=-17\degree$ & $6$ obstacles & easy \\
\hline
Test 2 & $x=3$, $z=0$,  $rot=0\degree$ & $17$ obstacles & medium \\
\hline
Test 3 & $x=-9$, $z=-25$, $rot=-21\degree$ & 5 obstacles & medium \\
\hline
Test 4 & $x=65$, $z=-45$, $rot=150\degree$ & 5 obstacles & hard \\
\hline
\end{tabular}
}
\label{table.environments}
\end{center}
\end{table*}  

\minisection{Test Environment 1} - the position of the building containing the goal is slightly different from the training environment, with a slight rotation to the left. Additionally, we introduce $5$ more obstacles in the path from the agent's starting position to the goal. We categorize the subjective difficulty of this variation as \textit{easy}, consistent with the training environment;

\minisection{Test Environment 2} - The position of the building containing the goal differs slightly along the x-axis, while the rotation remains the same as in the training environment. However, a total of $17$ obstacles obstruct the path from the agent's starting position to the goal. We categorize the subjective difficulty of this variation as \textit{medium};

\minisection{Test Environment 3} - The position of the building is significantly different from the one in the training environment, as can be seen by comparing Figure~\ref{fig:environments}(a) and (c). Furthermore, the building has a different rotation, with a higher degree compared to Test Environment 1. This environment also includes a total of $5$ obstacles. We categorize the subjective difficulty of this variation as \textit{medium}; and

\minisection{Test Environment 4} - The position of the building is entirely different from the training environment. In contrast to the training environment, where the goal is in front of the agent facing its direction, the goal in this test environment is behind the agent's starting position, facing its back. This configuration requires the agent to follow a completely different trajectory to reach the goal, one that is not in the expert dataset. Moreover, there are a total of $5$ obstacles. Due to the significant difference in the goal position, we categorize the subjective difficulty of this variation as \textit{hard}.

A summary of design changes and subjective complexity is available in Table~\ref{table.environments} and a visual for the modified environment is presented in Figure~\ref{fig:environments}(c).

\subsection{Augmented Models}
\label{section:augmented.data}
As a baseline, we used the dataset consisting of the original $78$ episodes of data without any applied augmentations. Using this data and the augmentations, we created different datasets, each featuring one or more augmentations. We refer to the models trained using these datasets as augmented models, and they are labeled with the corresponding dataset's augmentations. We applied all the augmentations listed in Section~\ref{augmentation.list} and all their combinations, up to $3$ augmentations, i.e, one dataset could be augmented with $1$ augmentation, $2$ augmentations, or $3$ augmentations. 
Preliminary experiments showed that combining Gaussian and uniform noise together decreased the overall performances. For this reason, we excluded the datasets that have both Gaussian and uniform noise. This process resulted in a total of $38$ different combinations of data augmentations.

\begin{figure}
  \centerline{\includegraphics[width=.7\columnwidth]{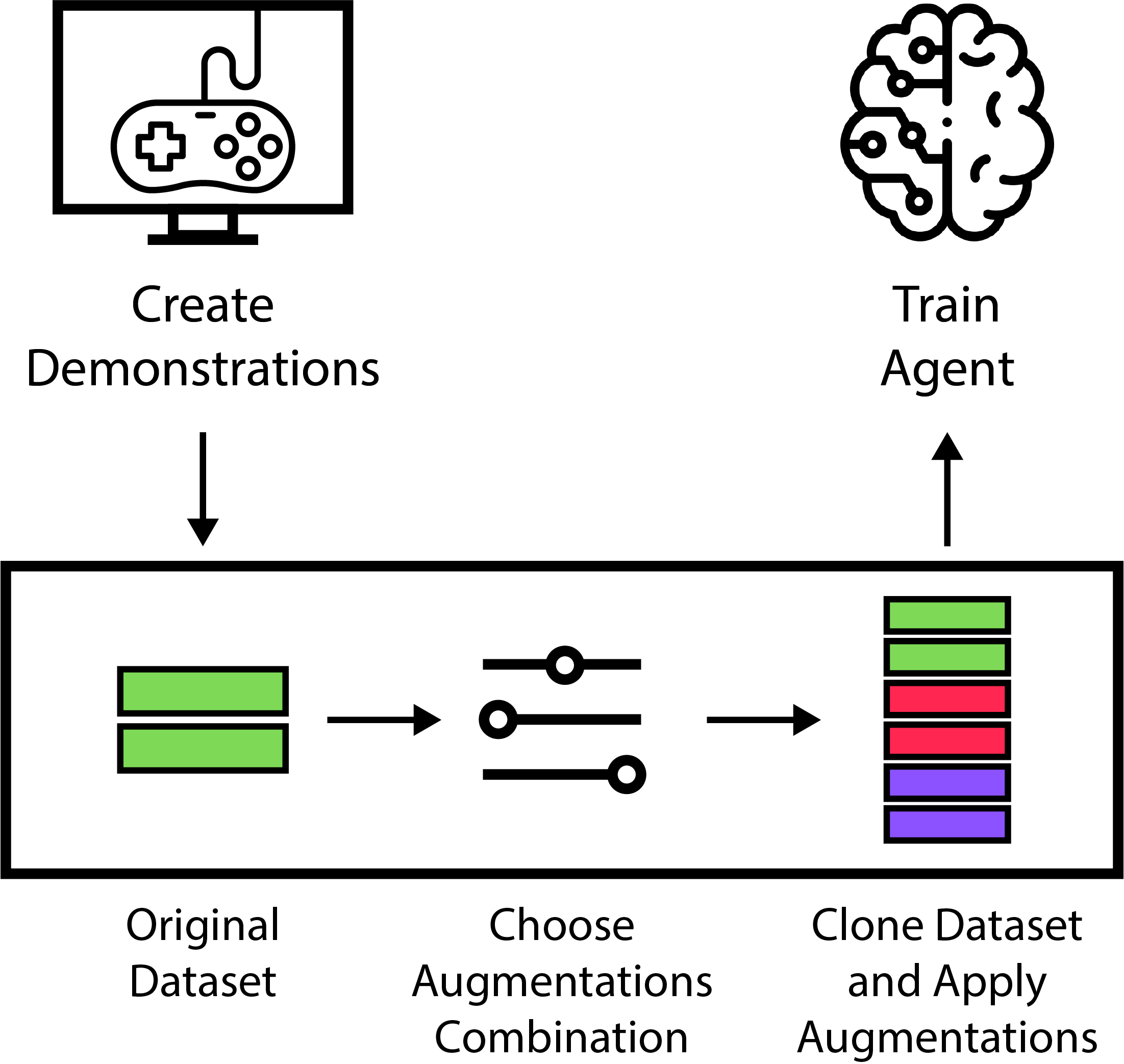}}
  \caption{Overview of the augmentation process. We start with the original set of demonstrations. Then, depending on the number and type of the selected augmentations, we create a new, unique dataset. For example, if we select two augmentations (Gaussian noise and semantic dropout), we start from the original dataset, apply Gaussian noise to it, and then augment the resulting dataset with semantic dropout.}
  \label{fig:aug-compisition}
\end{figure}

We initialized a unique training dataset for each model by starting with the original demonstration data. We then cloned the original data, sequentially applied each augmentation from the combination, and added the resulting data to the dataset. Additional details of how we applied the augmentations are provided in Section~\ref{sec:method}. We performed the augmentation step three times, resulting in a dataset four times as large as the original data. Augmentations were applied to the data in the order listed in \ref{augmentation.list}. This ordering was chosen to maximize the effect of changing the data by first performing translations followed by scaling, and to preserve dropout by performing those augmentations last in the composition. The process is outlined in Figure~\ref{fig:aug-compisition}. 

Dataset size and standard deviation of Gaussian noise are two significant parameters. To better understand their effects, we conduct a hyperparameter study over these parameters. The tested ranges were selected following preliminary experiments. All hyperparameters together with the variations are summarized in Table~\ref{table.hyperparameters}. With these variations, we trained a total of $228$ different models. Each model is a different combinations of augmentations, varying their hyperparameters. For instance, one model could be applying Gaussian noise with $\theta = 0.03$ plus state mixup to $50\%$ of the original data. As we previously said, we have a total of $38$ combinations times the $6$ different sizes of the dataset. For each model, we repeated the experiments for $10$ different seeds.

\begin{table*}
\caption{Summary of available hyperparameters. Values in square brackets were experimentally varied; all others remained fixed. The ranges were selected following preliminary experiments.}
\begin{center}
\scalebox{0.8}{
\begin{tabular}{|c|c|}
\hline
\textbf{\textit{Hyperparameter}}& \textbf{\textit{Values}}\\
\hline
Gaussian noise & $\mu=0, \sigma=[0.03, 0.003, 0.0003, 0.00003]$ \\
\hline
Data size & $[50\%, 60\%, 70\%, 80\%, 90\%, 100\%]$ \\
\hline
State mixup & $\alpha = 0.4, \beta=0.4$ \\
\hline
Uniform noise & $\text{low}=-0.0003, \text{high}=0.0003$ \\
\hline
Dropout, num. continuous features & $3$ \\
\hline
Dropout, num. semantic features & $12$ \\
\hline
Max. augmentation combinations & $3$ \\
\hline
Num. data clones & $3$ \\
\hline
\end{tabular}
}
\label{table.hyperparameters}
\end{center}
\end{table*}

\begin{table*}
\caption{Summary of average performance for models which outperformed the baseline. Models shown are taken from the set of $19$ top-performing models. Additional details of how the models were chosen are provided in Section~\ref{section:augmented.data}.}
\begin{center}
\scalebox{0.8}{
\begin{tabular}{|c|c|c|c|}
\hline
\textbf{\textit{Environment}}& \textbf{\textit{Mean relative success rate}}& \textbf{\textit{STD relative success rate}}& \textbf{\textit{Number of models which outperform the baseline in each environment}}\\
\hline
Train & $1.210187$ & $0.854240$ & $19$ \\
\hline
Test 1 & $1.765625$ & $1.214967$ & $16$\\
\hline
Test 2 & $1.197297$ & $0.976769$ & $5$ \\
\hline
Test 3 & $1.800574$ & $1.525612$ & $17$ \\
\hline
Test 4 & $1.234450$ & $1.028380$ & $1$ \\
\hline
\end{tabular}
}
\label{table.summary.top.basebeaters}
\end{center}
\end{table*}

\begin{figure}%
\centerline{\includegraphics[width=0.83\columnwidth]{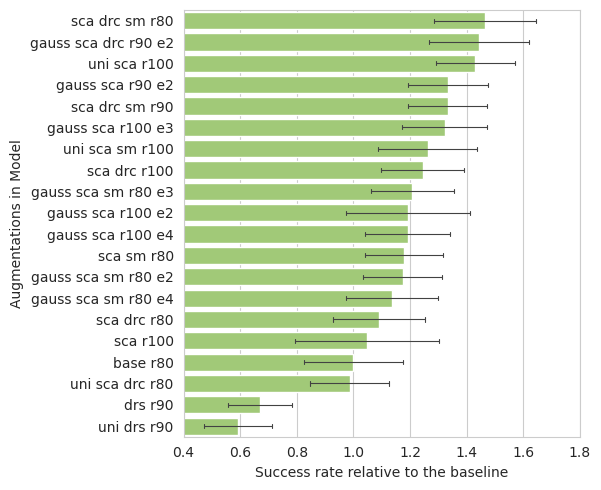}}
\caption{Relative performance of the top $20$ models, averaged over all experiments (i.e., the $4$ test environments) with standard error. Abbreviations: gauss (Gaussian noise), uni (uniform noise) sca (scaling), sm (state-mixup), drc (continuous dropout), drs (semantic dropout), rX (X percentage of the original dataset used), and eX ($\sigma = 3 \cdot 10^{-\text{X}}$). Additional details about the augmentations are provided in Section~\ref{sec:method}.}
\label{fig.top.avgmodelperformance}
\end{figure}

Figure~\ref{fig.top.avgmodelperformance} shows the relative performance of each of the top $19$ models, averaged over all experiments. The figure illustrates that augmentations can yield large improvements over the baseline model. However, the large standard deviation indicates that models may be sensitive to training parameters. In some instances, this can be interpreted as noise added to the data sending the agent into an unknown state, from which it cannot recover. We conduct further experiments to examine the standard deviation of the augmented models. Our findings suggest that the standard deviation of these models increases as the dataset size increases. This particular effect cannot be seen in the figure due to the fact that only a subset of  the models  are shown. For instance, the standard deviation of the model augmented with Gaussian noise ranges from 0.07 when using 50\% of the dataset to 1.11 when using 100\%. Similar trends were observed with all the other augmentations.

\begin{figure}%
\centerline{\includegraphics[width=0.9\columnwidth]{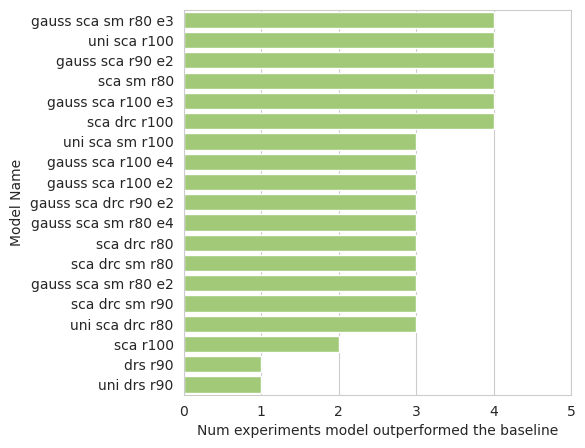}}
\caption{Models that consistently outperform the baseline in the $4$ different testing environments plus the training environment. For the abbreviations legend, see Figure~\ref{fig.top.avgmodelperformance}.}
\label{fig.top.consistency}
\end{figure}

\subsection{Evaluation}
\label{sec:evaluation}

During testing, agents run in each environment for $100$ episodes. Episodes are reset after a timeout of $750$ steps. The episode is recorded as a success if the agent reaches the goal before the timeout. For an initial evaluation, we run the full set of $228$ models in the initial training environment. Using the mean success rate over $10$ seeds of each model as a selection criterion, we narrow the set of models down to $40$ to evaluate generalization performance in the testing environments. This set of models consists of the following two groups: 1) the $19$ top-performing models and 2) a selection of $20$ of the lowest-performing models. For the lowest-performing group, we selected models with relative success rates ranging from $0.00$ to $0.39$ to obtain a mix of both low-performing and failing models. In addition, we selected the highest-performing baseline model and included it in the top-performing group. 
Our main focus is to evaluate how data augmentation can impact the generalization performance of the trained agents. For this reason, we evaluate only the top $19$ models found in the initial evaluation.
We are interested in finding the relative success rate of each augmented model compared to the baseline for each test environment. This provides a quantitative measure of the improvements, if any, that are achieved through the data augmentation combination, so we use this as our primary benchmark. We also seek to identify consistency in model performance across environments, as it might suggest a clear set of combinations to use---or to avoid---in similar scenarios, as a starting point for other game environments. 
% benchmark

Our first assessment is whether data augmentations could improve an imitation learning agent's ability to generalize to unseen changes in an environment. The results, summarized in Table~\ref{table.summary.top.basebeaters}, show that these augmented models outperform the baseline in each of the test environments. 
Note that a mean relative success rate of 1.X corresponds to an improvement of X\%.
Hence, the mean relative success rates showed an improvement of $20\%$ up to $80\%$ for different environments but with a relatively high variance over different models.

The next assessment is to determine whether a combination of augmentations consistently improves generalization regardless of the environment. Figure~\ref{fig.top.consistency} shows that $6$ different models outperform their respective baselines in $4$ of the $5$ environments. These $6$ models each includes a combination of at least $2$ augmentations and use at least $80\%$ of the data. 
While these models are the most consistent in outperforming the baseline, none of them have the highest relative success rates, but $3$ of them are in the top $6$ in the same evaluation (see Figure~\ref{fig.top.avgmodelperformance}). This suggests that there is a trade-off between best achievable generalization performance and consistency over all testing environments.

\begin{figure}%
\centerline{\includegraphics[width=0.8\columnwidth]{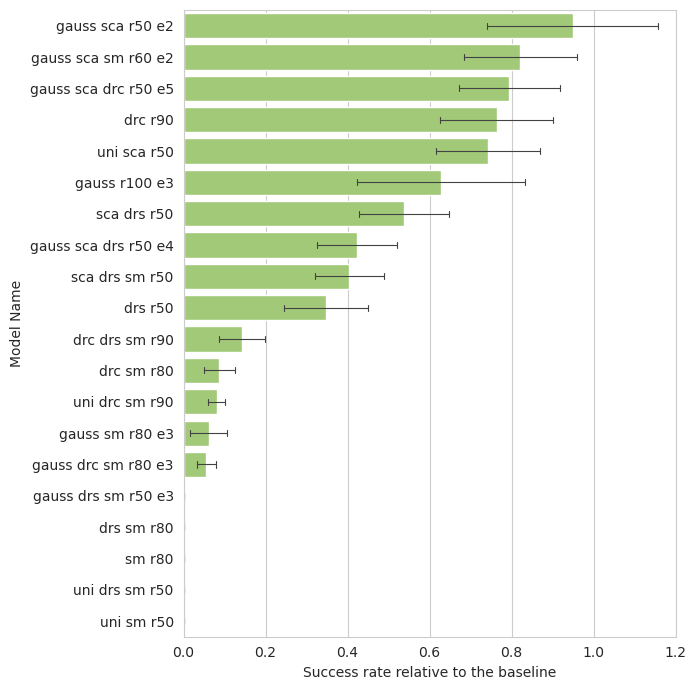}}
\caption{Relative performance, averaged over all environments, is consistently poor or failing for the bottom group of models. Abbreviations: gauss (Gaussian noise), uni (uniform noise) sca (scaling), sm (state-mixup), drc (continuous dropout), drs (semantic dropout), rX (X percentage of the original dataset used), and eX ($\sigma = 3 \cdot 10^{-\text{X}}$).}
\label{fig.bottom.consistency}
\end{figure}

For the $20$ lowest-performing models in the training environment defined above, we assess whether their poor relative performance is consistent between the training environment and across the testing environments. Figure~\ref{fig.bottom.consistency} shows the average success rate relative to the baseline across all environments remains quite low and these results are consistent with the results from the training environment.

Finally, this study aims to determine if there is a single most effective augmentation for agent performance and if it exists, to identify it. To do this, we grouped all models tested in the testing environment by augmentation, and computed the average success rate relative to the baseline. For example, in the Gaussian noise group, we include all models containing \textit{gauss} in their set of augmentations. From this analysis, we found the following average relative success rates: scaling ($1.27$), state mixup ($1.26$), continuous dropout ($1.26$), Gaussian noise ($1.25$), uniform noise ($1.02$), and semantic dropout ($0.50$). 
These investigations, combined with the consistency study, highlight scaling as one of the most promising augmentations. State mixup is one of the most effective augmentations for generalization in certain environments but it did not outperform the baseline as consistently as models augmented with scaling. Gaussian noise and continuous dropout follow the same trend, with the latter exhibiting the lowest consistency. Lastly, uniform noise had the least impact on generalization while semantic dropout clearly had a negative effect on generalization. Nevertheless, due to the sensitivity to the training parameters previously mentioned, particularly for noise-based augmentations, these results should be considered as promising starting points for further studies to cover a wider range of the possible augmentations with different parameters.

\section{Conclusion and discussions}
In this paper we have conducted a comprehensive study of data augmentation in imitation learning. Training self-learning agents with either reinforcement learning or imitation learning, involves a generalization--efficiency trade-off: we can increase generalization, but at the cost of sample efficiency. Data augmentation can help improve generalization while maintaining sample efficiency. Building on the success of data augmentation in supervised learning and reinforcement learning, we investigated data augmentation techniques for feature-based imitation learning and their effects on the training of imitative agents. We evaluated our agents in four distinct 3D test environments.
The main findings of this paper are twofold. First, data augmentations can indeed improve performance in an imitation learning setting. The imitative agents demonstrated improved generalization in previously unseen situations when trained on an augmented dataset. Second, our study indicates that certain combinations of data augmentations consistently enhance performance across a variety of generalization environments, even if the performance of individual augmentations may vary depending on the parameters used.

This improvement in generalization with augmented data is not surprising. Data augmentation in supervised learning involves manipulating the images to resemble a visual system. Rotation, cropping, translation, noise, etc., are augmentations that can be translated into game-state observations. The challenge lies in identifying which augmentations contribute to generalization and which do not. Our preliminary study revealed that combining multiple augmentations yields better results than using single augmentations alone. At the same time, of the single augmentations tested, scaling emerged as one of the most consistently performing throughout our study, followed by continuous dropout and Gaussian noise.

This study provides a promising research direction for data augmentation and imitation learning. However, several limitations warrant consideration. Our preliminary results indicate that certain augmentations are particularly sensitive to the training parameters, thus requiring a more thorough study to assess their contribution. Furthermore, we tested these augmentations only in an internal game environment, and exclusively on navigation and interaction tasks. Exploring these augmentations in different environments, and with different tasks, should prove valuable. Furthermore, although most game AI agents use a similar state space, alternative solutions exist. Hence, testing these augmentations in a larger range of environments, task and observation spaces is a promising future research direction. 
% slightly different observation space

{\footnotesize \bibliography{biblio}}

% Generated by IEEEtranN.bst, version: 1.14 (2015/08/26)
\begin{thebibliography}{34}
\providecommand{\natexlab}[1]{#1}
\providecommand{\url}[1]{#1}
\csname url@samestyle\endcsname
\providecommand{\newblock}{\relax}
\providecommand{\bibinfo}[2]{#2}
\providecommand{\BIBentrySTDinterwordspacing}{\spaceskip=0pt\relax}
\providecommand{\BIBentryALTinterwordstretchfactor}{4}
\providecommand{\BIBentryALTinterwordspacing}{\spaceskip=\fontdimen2\font plus
\BIBentryALTinterwordstretchfactor\fontdimen3\font minus \fontdimen4\font\relax}
\providecommand{\BIBforeignlanguage}[2]{{%
\expandafter\ifx\csname l@#1\endcsname\relax
\typeout{** WARNING: IEEEtranN.bst: No hyphenation pattern has been}%
\typeout{** loaded for the language `#1'. Using the pattern for}%
\typeout{** the default language instead.}%
\else
\language=\csname l@#1\endcsname
\fi
#2}}
\providecommand{\BIBdecl}{\relax}
\BIBdecl

\bibitem[Stahlke et~al.(2019)Stahlke, Nova, and Mirza-Babaei]{playfulness}
S.~Stahlke, A.~Nova, and P.~Mirza-Babaei, ``Artificial playfulness: A tool for automated agent-based playtesting,'' in \emph{CHI Conference on Human Factors in Computing Systems}, 2019.

\bibitem[Cho et~al.(2010)Cho, Sohn, Park, and Kang]{scenariobased}
C.-S. Cho, K.-M. Sohn, C.-J. Park, and J.-H. Kang, ``Online game testing using scenario-based control of massive virtual users,'' in \emph{International Conference on Advanced Communication Technology}, 2010.

\bibitem[Vinyals et~al.(2019)Vinyals, Babuschkin, Czarnecki, Mathieu, Dudzik, Chung, Choi, Powell, Ewalds, Georgiev, et~al.]{alphastar}
O.~Vinyals, I.~Babuschkin, W.~M. Czarnecki, M.~Mathieu, A.~Dudzik, J.~Chung, D.~H. Choi, R.~Powell, T.~Ewalds, P.~Georgiev \emph{et~al.}, ``Grandmaster level in {S}tar{C}raft {II} using multi-agent reinforcement learning,'' \emph{Nature}, vol. 575, no. 7782, pp. 350--354, 2019.

\bibitem[OpenAI et~al.(2019)OpenAI, Berner, Brockman, Chan, Cheung, Debiak, Dennison, Farhi, Fischer, Hashme, Hesse, Józefowicz, Gray, Olsson, Pachocki, , et~al.]{openai}
OpenAI, C.~Berner, G.~Brockman, B.~Chan, V.~Cheung, P.~Debiak, C.~Dennison, D.~Farhi, Q.~Fischer, S.~Hashme, C.~Hesse, R.~Józefowicz, S.~Gray, C.~Olsson, J.~Pachocki,  \emph{et~al.}, ``Dota 2 with large scale deep reinforcement learning,'' \emph{arXiv preprint 1912.06680}, 2019.

\bibitem[Wurman et~al.(2022)Wurman, Barrett, Kawamoto, MacGlashan, Subramanian, Walsh, Capobianco, Devlic, Eckert, Fuchs, et~al.]{gtsophy}
P.~R. Wurman, S.~Barrett, K.~Kawamoto, J.~MacGlashan, K.~Subramanian, T.~J. Walsh, R.~Capobianco, A.~Devlic, F.~Eckert, F.~Fuchs \emph{et~al.}, ``Outracing champion gran turismo drivers with deep reinforcement learning,'' \emph{Nature}, vol. 602, no. 7896, pp. 223--228, 2022.

\bibitem[Bergdahl et~al.(2020)Bergdahl, Gordillo, Tollmar, and Gissl{\'e}n]{augmenting}
J.~Bergdahl, C.~Gordillo, K.~Tollmar, and L.~Gissl{\'e}n, ``Augmenting automated game testing with deep reinforcement learning,'' in \emph{Conference on Games (CoG)}, 2020.

\bibitem[Gordillo et~al.(2021)Gordillo, Bergdahl, Tollmar, and Gisslén]{improving}
C.~Gordillo, J.~Bergdahl, K.~Tollmar, and L.~Gisslén, ``Improving playtesting coverage via curiosity driven reinforcement learning agents,'' in \emph{Conference on Games (CoG)}, 2021.

\bibitem[Sestini et~al.(2022{\natexlab{a}})Sestini, Bergdahl, Tollmar, Bagdanov, and Gissl{\'e}n]{sestini2022kiwi}
A.~Sestini, J.~Bergdahl, K.~Tollmar, A.~D. Bagdanov, and L.~Gissl{\'e}n, ``Towards informed design and validation assistance in computer games using imitation learning,'' in \emph{NeurIPS workshop on Human In the Loop Learning}, 2022.

\bibitem[Justesen et~al.(2018)Justesen, Torrado, Bontrager, Khalifa, Togelius, and Risi]{illuminating}
N.~Justesen, R.~Torrado, P.~Bontrager, A.~Khalifa, J.~Togelius, and S.~Risi, ``\BIBforeignlanguage{English}{Illuminating generalization in deep reinforcement learning through procedural level generation},'' \emph{\BIBforeignlanguage{English}{NIPS Workshop on DRL}}, 2018.

\bibitem[Sestini et~al.(2022{\natexlab{b}})Sestini, Gissl{\'e}n, Bergdahl, Tollmar, and Bagdanov]{sestini2022ccpt}
A.~Sestini, L.~Gissl{\'e}n, J.~Bergdahl, K.~Tollmar, and A.~D. Bagdanov, ``Automated gameplay testing and validation with curiosity-conditioned proximal trajectories,'' \emph{IEEE Transactions on Games}, 2022.

\bibitem[He et~al.(2020)He, Fan, Wu, Xie, and Girshick]{supervisedaug1}
K.~He, H.~Fan, Y.~Wu, S.~Xie, and R.~Girshick, ``Momentum contrast for unsupervised visual representation learning,'' in \emph{conference on computer vision and pattern recognition}, 2020, pp. 9729--9738.

\bibitem[Xie et~al.(2020)Xie, Dai, Hovy, Luong, and Le]{supervisedaug2}
Q.~Xie, Z.~Dai, E.~Hovy, T.~Luong, and Q.~Le, ``Unsupervised data augmentation for consistency training,'' \emph{Advances in neural information processing systems}, vol.~33, pp. 6256--6268, 2020.

\bibitem[Yarats et~al.(2021)Yarats, Fergus, Lazaric, and Pinto]{drq}
D.~Yarats, R.~Fergus, A.~Lazaric, and L.~Pinto, ``Mastering visual continuous control: Improved data-augmented reinforcement learning,'' \emph{arXiv preprint arXiv:2107.09645}, 2021.

\bibitem[Sinha et~al.(2022)Sinha, Mandlekar, and Garg]{s4rl}
S.~Sinha, A.~Mandlekar, and A.~Garg, ``S4rl: Surprisingly simple self-supervision for offline reinforcement learning in robotics,'' in \emph{Conference on Robot Learning}.\hskip 1em plus 0.5em minus 0.4em\relax PMLR, 2022, pp. 907--917.

\bibitem[Cobbe et~al.(2020)Cobbe, Hesse, Hilton, and Schulman]{procgen}
K.~Cobbe, C.~Hesse, J.~Hilton, and J.~Schulman, ``Leveraging procedural generation to benchmark reinforcement learning,'' in \emph{International conference on machine learning}.\hskip 1em plus 0.5em minus 0.4em\relax PMLR, 2020, pp. 2048--2056.

\bibitem[Bain and Sammut(1995)]{bc1}
M.~Bain and C.~Sammut, ``A framework for behavioural cloning.'' in \emph{Machine Intelligence 15}, 1995, pp. 103--129.

\bibitem[Ross et~al.(2011)Ross, Gordon, and Bagnell]{dagger}
S.~Ross, G.~Gordon, and D.~Bagnell, ``A reduction of imitation learning and structured prediction to no-regret online learning,'' in \emph{2011 International Conference on Artificial Intelligence and Statistics {(ICAIS)}}, 2011.

\bibitem[Ho and Ermon(2016)]{gail}
J.~Ho and S.~Ermon, ``Generative adversarial imitation learning,'' in \emph{International Conference on Neural Information Processing Systems}, 2016.

\bibitem[Pearce and Zhu(2022)]{ilcounter}
T.~Pearce and J.~Zhu, ``Counter-strike deathmatch with large-scale behavioural cloning,'' in \emph{2022 IEEE Conference on Games (CoG)}.\hskip 1em plus 0.5em minus 0.4em\relax IEEE, 2022, pp. 104--111.

\bibitem[Amiranashvili et~al.(2020)Amiranashvili, Dorka, Burgard, Koltun, and Brox]{minecraft}
A.~Amiranashvili, N.~Dorka, W.~Burgard, V.~Koltun, and T.~Brox, ``Scaling imitation learning in minecraft,'' \emph{arXiv preprint arXiv:2007.02701}, 2020.

\bibitem[Chang et~al.(2019)Chang, Aytemiz, and Smith]{reveal}
K.~Chang, B.~Aytemiz, and A.~M. Smith, ``Reveal-more: Amplifying human effort in quality assurance testing using automated exploration,'' in \emph{2019 Conference on Games (CoG)}.\hskip 1em plus 0.5em minus 0.4em\relax IEEE, 2019, pp. 1--8.

\bibitem[Zhao et~al.(2020)Zhao, Borovikov, Silva, Beirami, Rupert, Somers, Harder, Kolen, Pinto, Pourabolghasem, et~al.]{winning}
Y.~Zhao, I.~Borovikov, F.~D.~M. Silva, A.~Beirami, J.~Rupert, C.~Somers, J.~Harder, J.~Kolen, J.~Pinto, R.~Pourabolghasem \emph{et~al.}, ``Winning isn't everything: Enhancing game development with intelligent agents,'' \emph{Transactions on Games}, 2020.

\bibitem[Harmer et~al.(2018)Harmer, Gissl{\'e}n, del Val, Holst, Bergdahl, Olsson, Sj{\"o}{\"o}, and Nordin]{concurrent}
J.~Harmer, L.~Gissl{\'e}n, J.~del Val, H.~Holst, J.~Bergdahl, T.~Olsson, K.~Sj{\"o}{\"o}, and M.~Nordin, ``Imitation learning with concurrent actions in 3d games,'' in \emph{Conference on Computational Intelligence and Games}.\hskip 1em plus 0.5em minus 0.4em\relax IEEE, 2018.

\bibitem[Tucker et~al.(2018)Tucker, Gleave, and Russell]{adam}
A.~Tucker, A.~Gleave, and S.~Russell, ``Inverse reinforcement learning for video games,'' in \emph{NIPS Workshop on Deep Reinforcement Learning}, 2018.

\bibitem[Bellemare et~al.(2013)Bellemare, Naddaf, Veness, and Bowling]{ale}
M.~G. Bellemare, Y.~Naddaf, J.~Veness, and M.~Bowling, ``The arcade learning environment: An evaluation platform for general agents,'' \emph{Journal of Artificial Intelligence Research}, vol.~47, pp. 253--279, 2013.

\bibitem[Ferguson et~al.(2022)Ferguson, Devlin, Kudenko, and Walker]{ferguson2022imitating}
M.~Ferguson, S.~Devlin, D.~Kudenko, and J.~A. Walker, ``Imitating playstyle with dynamic time warping imitation,'' in \emph{International Conference on the Foundations of Digital Games}, 2022, pp. 1--11.

\bibitem[Pearce et~al.(2023)Pearce, Rashid, Kanervisto, Bignell, Sun, Georgescu, Macua, Tan, Momennejad, Hofmann, et~al.]{imitating}
T.~Pearce, T.~Rashid, A.~Kanervisto, D.~Bignell, M.~Sun, R.~Georgescu, S.~V. Macua, S.~Z. Tan, I.~Momennejad, K.~Hofmann \emph{et~al.}, ``Imitating human behaviour with diffusion models,'' \emph{arXiv preprint arXiv:2301.10677}, 2023.

\bibitem[Sestini et~al.(2021)Sestini, Kuhnle, and Bagdanov]{deairl}
A.~Sestini, A.~Kuhnle, and A.~D. Bagdanov, ``Demonstration-efficient inverse reinforcement learning in procedurally generated environments,'' in \emph{2021 Conference on Games (CoG)}.\hskip 1em plus 0.5em minus 0.4em\relax IEEE, 2021, pp. 1--8.

\bibitem[Sestini et~al.(2020)Sestini, Kuhnle, and Bagdanov]{deepcrawl}
------, ``Deepcrawl: Deep reinforcement learning for turn-based strategy games,'' \emph{arXiv preprint arXiv:2012.01914}, 2020.

\bibitem[Romac et~al.(2021)Romac, Portelas, Hofmann, and Oudeyer]{teachmyagent}
C.~Romac, R.~Portelas, K.~Hofmann, and P.-Y. Oudeyer, ``Teachmyagent: a benchmark for automatic curriculum learning in deep rl,'' in \emph{International Conference on Machine Learning}.\hskip 1em plus 0.5em minus 0.4em\relax PMLR, 2021, pp. 9052--9063.

\bibitem[Mumuni and Mumuni(2022)]{survey}
A.~Mumuni and F.~Mumuni, ``Data augmentation: A comprehensive survey of modern approaches,'' \emph{Array}, p. 100258, 2022.

\bibitem[Laskin et~al.(2020)Laskin, Lee, Stooke, Pinto, Abbeel, and Srinivas]{rlaugmentation}
M.~Laskin, K.~Lee, A.~Stooke, L.~Pinto, P.~Abbeel, and A.~Srinivas, ``Reinforcement learning with augmented data,'' \emph{Advances in neural information processing systems}, vol.~33, pp. 19\,884--19\,895, 2020.

\bibitem[Zolna et~al.(2021)Zolna, Reed, Novikov, Colmenarejo, Budden, Cabi, Denil, de~Freitas, and Wang]{augmentationil}
K.~Zolna, S.~Reed, A.~Novikov, S.~G. Colmenarejo, D.~Budden, S.~Cabi, M.~Denil, N.~de~Freitas, and Z.~Wang, ``Task-relevant adversarial imitation learning,'' in \emph{Conference on Robot Learning}, 2021.

\bibitem[Zhang et~al.(2017)Zhang, Cisse, Dauphin, and Lopez-Paz]{zhang2017mixup}
H.~Zhang, M.~Cisse, Y.~N. Dauphin, and D.~Lopez-Paz, ``mixup: Beyond empirical risk minimization,'' \emph{arXiv preprint arXiv:1710.09412}, 2017.

\end{thebibliography}
\bibliographystyle{IEEEtranN}
\end{document}